\documentclass[conference]{IEEEtran}  

\IEEEoverridecommandlockouts                              

\usepackage{amsmath} 
\usepackage{amssymb}  
\usepackage{graphicx}
\usepackage{subfigure}
\usepackage{multirow}
\usepackage{color}
\usepackage{hyperref}
\usepackage{graphicx}
\usepackage{float}
\usepackage{caption}
\usepackage{svg}
\usepackage{wrapfig}
\usepackage{xurl}
\usepackage{hyperref}
\usepackage{multicol}
\usepackage{amsmath}
\usepackage{amssymb}
\usepackage{eucal}
\usepackage{algorithm}
\usepackage{algpseudocode}
\usepackage{tabularx}
\usepackage{booktabs}
\usepackage{multirow}
\usepackage{xcolor}
\usepackage{lipsum}

\usepackage{tikz}
\usepackage{textcomp}
\usepackage{collcell}
\usepackage{color, xcolor, colortbl}
\usepackage{arydshln}
\usepackage{enumerate}%

\IEEEoverridecommandlockouts                              

\title{\LARGE \bf
Instance-wise Grasp Synthesis for Robotic Grasping}

\author{Yucheng Xu$^{1}$, Mohammadreza Kasaei$^{1}$, Hamidreza Kasaei$^{2}$, and Zhibin Li$^{3}$
\thanks{$^{1}$ Yucheng Xu and Mohammadreza Kasaei are with the School of Informatics, University of Edinburgh, UK. Email: \{yucheng.xu, m.kasaei\}@ed.ac.uk}%
\thanks{$^{2}$Hamidreza~Kasaei is with the Department of Artificial Intelligence, Bernoulli Institute, Faculty of Science and Engineering, University of Groningen, The Netherlands. Email: hamidreza.kasaei@rug.nl}
\thanks{$^{3}$ Zhibin~Li is with the Department of Computer Science, University College London, UK. Email: alex.li@ucl.ac.uk}%
\thanks{This work is supported by EU H2020 project Enhancing Healthcare with Assistive Robotic Mobile Manipulation (HARMONY, 101017008).} }

\begin{document}

\maketitle
\thispagestyle{empty}
\pagestyle{empty}

\begin{abstract}

Generating high-quality instance-wise grasp configurations provides critical information of how to grasp specific objects in a multi-object environment and is of high importance for robot manipulation tasks. This work proposed a novel \textbf{S}ingle-\textbf{S}tage \textbf{G}rasp (SSG) synthesis network, which performs high-quality instance-wise grasp synthesis in a single stage: instance mask and grasp configurations are generated for each object simultaneously. Our method outperforms state-of-the-art on robotic grasp prediction based on the OCID-Grasp dataset, and performs competitively on the JACQUARD dataset. The benchmarking results showed significant improvements compared to the baseline on the accuracy of generated grasp configurations. The performance of the proposed method has been validated through both extensive simulations and real robot experiments for three tasks including single object pick-and-place, grasp synthesis in cluttered environments and table cleaning task.

\end{abstract}

\section{INTRODUCTION}

In human-centered environments, robots are becoming increasingly useful in a variety of applications related to manipulating specific objects, thus a robust and efficient instance-wise grasp synthesis approach is of great importance, as it provides vital information~(e.g., location and grasp configurations) for manipulating target objects. Image-based instance-wise grasp synthesis in cluttered environments is yet a very challenging task. It aims at generating high-quality grasp configurations for specific objects in the multi-object scenario using a single image as the input. In this paper, we seek to leverage the success of prior research on semantic instance segmentation as well as generative grasp synthesis to design a novel model, which  solves instance-wise grasp synthesis tasks in a single-stage manner for robotic manipulations.

Designing an image-based instance-wise grasp synthesis model is difficult for two key reasons: (i)~current 2D grasp synthesis approaches either employ a region proposal network to find graspable regions~\cite{8793751, SONG2020101963, zhang2019roi, luo2020grasp}, or adopt generative model to predict pixel-wise grasp configurations~\cite{morrison2018closing, kumra2020antipodal, cao2021lightweight, li2022novel}. Both of these approaches are limited to scene-level grasp synthesis; in other words, they can only determine which parts of the scene are graspable, but not which objects. (ii) Since the grasp configurations are mostly generated from regional or global features, the relationship between objects and grasps is not clear. Thus, it is difficult to determine the grasp affiliations.

Recent research tackle instance-wise grasp synthesis tasks in a two-stage way~\cite{zhang2019multi, yang2019task, ainetter2021end, li2021keypoint}: (i)~in the first stage, grasp configurations will be generated for all graspable regions of the global input; (ii) then, the generated grasp configurations will be assigned to specific objects with the help of additional information, which is often derived from object detection or semantic segmentation. Two-stage methods inherently lack the relationship between predicted grasp configurations and detected objects, since the object detection task and object detection/segmentation task are completed separately. These methods mostly suffer from inaccurate grasp assignment, lack of class-specific information, and inefficiency stemmed from its cascade structure~\cite{zhang2019multi, li2021keypoint} (Details in Fig.~\ref{fig:failure-cases}, Section~\ref{sec:eval-ocid}).  

To address these limitations and solve instance grasp synthesis tasks in a more efficient and accurate way, we proposed the \textbf{S}ingle-\textbf{S}tage \textbf{G}rasp (SSG) synthesis model. The term ``single-stage" stands for the way of generating instance-wise grasp configurations. The grasp configurations are generated for each object instance directly without additional refinement or assignment modules which are commonly used in previous methods~\cite{zhang2019multi, yang2019task, ainetter2021end, li2021keypoint}. 

\begin{figure*}[!t]
\centering
\includegraphics[trim=2.5mm 2mm 2mm 2mm,clip,width=0.8\linewidth]{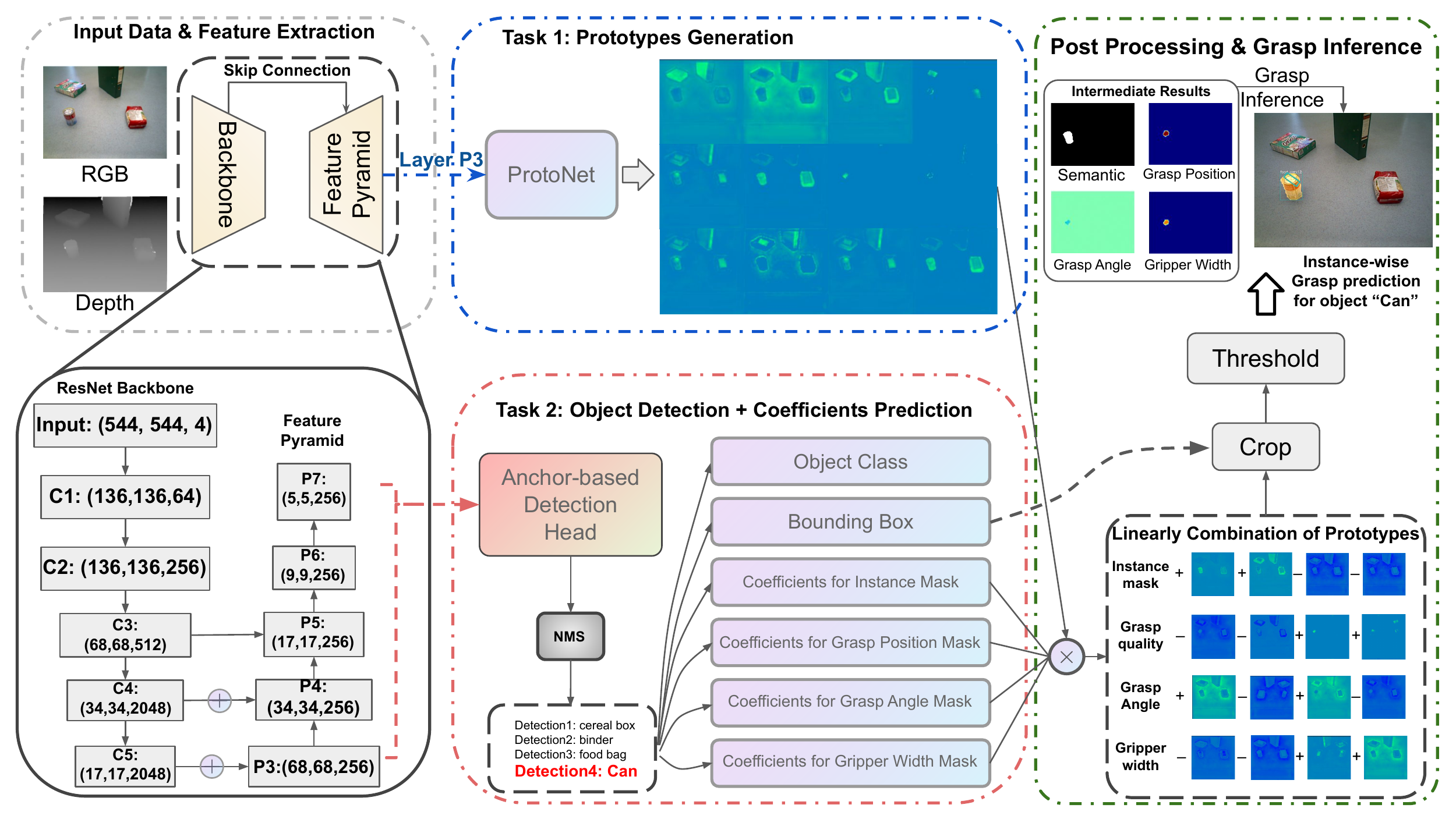}
\caption{System structure of the proposed model. The grasp configurations is generated as follows: (1) Feature extraction; (2) Generating of prototype masks; (3) Object detection, prediction of coefficients; (4) Linearly combination of prototypes with different predicted coefficients to generate instance mask and grasp masks; (5) Post processing to infer grasp configurations from generated grasp masks.} 
\vspace{-6mm}
\label{fig:ssg}
\end{figure*}

Our proposed SSG formulates the instance-wise grasp synthesis as two parallel tasks. The first task focuses on generating a set of prototype masks for the input RGB-D image, which can be regarded as vocabulary or global descriptors. The second task is to detect objects in the image and predicts extra sets of coefficients for each detected object. Finally, for each object that survives Non-Maximum Suppression~(NMS), those sets of coefficients are used to linearly assemble prototype masks to generate both instance segmentation and grasp masks. Here, grasp masks refer to pixel-wise grasp configurations proposed by~\cite{morrison2018closing, kumra2020antipodal}. In our proposed method, SSG, bounding box, class label, instance mask, and grasp masks are generated in parallel for each detected object which strongly keep the relationship between objects and grasps. The overall architecture (Fig.~\ref{fig:ssg}) clearly delineates the unique process of the proposed method.

The contributions are summarized as follows: \textbf{(1)} A novel \textbf{S}ingle-\textbf{S}tage \textbf{G}rasp (SSG) synthesis model, which solves the challenging instance-wise grasp synthesis tasks in a single-stage manner; \textbf{(2)} The proposed SSG outperforms state-of-the-art on OCID-Grasp dataset, and performs competitively on JACQUARD dataset, through the evaluation on extensive tests and validations in both simulations and real robot experiments.

The proposed SSG succeeded in synthesizing instance-wise grasp configurations in highly cluttered scenarios, where objects had 10\% to 25\% of overlapped areas, while other two-stage methods failed due to the mismatch between grasps and objects, and segmentation errors. Further, we demonstrate the scalability of the proposed method by extending it to affordance detection tasks~(See Section~\ref{sec:scalability} in details), and show the proposed method can be used as a general pipeline for multiple robot manipulation tasks.

\section{Related Works}
\label{sec:related}

Learning-based 2D robotic grasp synthesis has been increasingly attracting attention in past years~\cite{bohg2013data}. Modern learning-based 2D grasp synthesis approaches can be roughly categorized into detection-based and generative approaches. Detection-based approaches adopt object detection pipelines and treat grasp synthesis task as a detection task, since grasp configurations can be represented as  rotated rectangles in image plane~\cite{8793751, SONG2020101963,zhang2019roi, lenz2015deep, redmon2015real}. The work of~\cite{8793751} performed transfer learning between object detection and grasp detection. A Rotated Region Proposal Network~(RRPN), which is pre-trained on object detection dataset, is adopted to generate graspable region proposals. A single-stage grasp detection network purely based on Region Proposal Network~(RPN) was proposed in~\cite{SONG2020101963}. The grasp rectangles are directly regressed and classified from oriented anchors which are generated from RPN. ROI-GD~\cite{zhang2019roi} is a two-stage approach that detects grasp synthesis for specific regions by leveraging features from the object region rather than global input.

On the other hand, generative approaches produce pixel-wise grasp configurations for an input image~\cite{morrison2018closing, kumra2020antipodal, cao2021lightweight, li2022novel}. In this category, GG-CNN~\cite{morrison2018closing} approach aims to predict pixel-wise grasp configurations from depth images using generative neural network, where grasp configurations are embedded into three target masks representing grasp quality, grasp angle, and width of gripper respectively. Based on such representation of grasp configurations, the work in~\cite{kumra2020antipodal} introduced residual structure into generative neural network. Also, Guassian kernel are introduced in~\cite{cao2021lightweight, li2022novel, cheng2022robot} to better represent grasp configurations. In comparison with detection-based grasp synthesis methods, generative grasp synthesis methods avoid the generation of redundant region proposals and discrete sampling.

Despite improvements in learning based grasp synthesis~\cite{zhang2019roi, kumra2020antipodal}, instance-wise grasp synthesis is still challenging. Most approaches solve instance-wise grasp prediction problems indirectly, by defining a set of surrogate detection and assignment tasks~\cite{zhang2019multi, yang2019task, ainetter2021end, li2021keypoint}. In such pipelines, additional semantic segmentation or object detection branches are commonly adopted to assign grasp candidates to a specific object. 

Representative multi-task frameworks were proposed in ~\cite{zhang2019multi, li2021keypoint} which include two networks for object detection and grasp detection respectively. The results of object detection were adopted to assign grasp candidates to specific objects. TOG-CRFs proposed in~\cite{yang2019task} adds a Conditional Random Field~(CRF) to the grasp detection network, which models semantic contents of object regions to enable task-oriented grasp synthesis. Another work in~\cite{ainetter2021end} adds an semantic segmentation branch alongside the grasp detection branch to refine grasp candidates and assign them to target objects. 

Mask-Grasp RCNN proposed in~\cite{kamel2021mask} is based on Mask-RCNN~\cite{he2017mask}: a instance segmentation network. The method in~\cite{kamel2021mask} adds additional regression heads to the Mask-RCNN~\cite{he2017mask} to detect and regress grasps from aligned Regions-of-Interest~(RoIs) directly for each detected object instance. Mask-Grasp RCNN~\cite{kamel2021mask} is the first single-stage instance-wise grasp synthesis method, which is used in this work as a baseline of a single-stage method for the comparison study.

\section{Proposed Method}
\label{sec:method}

\subsection{Problem Formulation}
\label{sec:method:prob}
This work aims to synthesize grasp configurations for each object from an RGB-D image in a single-stage way. The task is defined as: to predict grasp configurations for each detected instance in an image plane. Grasp configurations in an image plane are commonly represented as rotated rectangle:~$\text{Grasp}_{\text{rect}} = (x, y, \theta, w, q)$, such as the formulation in ~\cite{8793751, SONG2020101963, zhang2019roi,lenz2015deep, redmon2015real}, where $(x, y)$ corresponds to the center of grasp rectangle in the image coordinates, $\theta$ is the rotation in camera's frame of reference, $w$ is the required width of gripper, $q$ is the quality of grasping. In our method, we formulated an additional label~($cls$) for each grasp configurations that indicate which object it belongs to.

\begin{figure}[t]
  \centering
  \vspace{-2mm}
  \includegraphics[trim=0 10.3cm 0 0,clip,width=\linewidth]{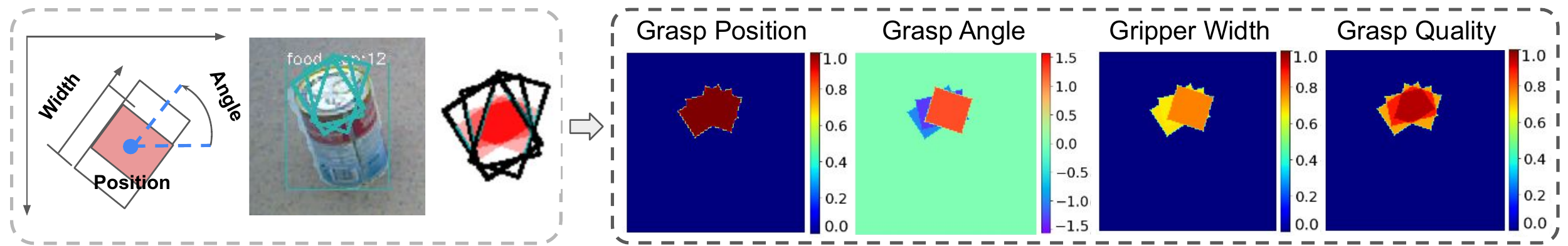}
  \caption{2D grasp rectangles are embedded into four different masks to represent grasp quality, grasp position, grasp angle and gripper width.}
  \label{fig:grasp-embed}
  \vspace{-6mm}
\end{figure}

To enable instance-wise grasp prediction, we adopted an approach similar to~\cite{morrison2018closing, kumra2020antipodal}, and further developed a grasp representation that can be integrated with the existing instance segmentation framework. For each object instance, we embed its ground truth grasp configurations into multiple masks indicating grasp position, grasp quality, grasp angle and gripper width~(see Fig.~\ref{fig:grasp-embed}). For a better representation of grasp quality, for each pixel, we calculate the number of overlapped grasp rectangles which include the pixel itself, and use $Sigmoid$ function to generate the grasp quality mask.

\subsection{Architecture}
\label{sec:method:architecture}
We developed the SSG, a single-stage grasp synthesis model, with insights from YOLACT~\cite{bolya2019yolact}. Fig.~\ref{fig:ssg} details the sub-modules and the workflow of our proposed method. First, a feature extraction module, consisting of ResNet-101~\cite{he2016deep} and Feature Pyramid Network~(FPN)~\cite{lin2017feature}, is adopted to extract shared multi-scale features from input RGB-D image. Deep layers~(C3, C4, C5) from ResNet-101 module are linked to FPN. Then, the ProtoNet branch, which is a fully convolutional network~\cite{long2015fully} with $k$-channel output, is used to generate a set of 32 prototype masks~($k=32$) for the entire input RGB-D image. P3 layer of FPN is used as the input of ProtoNet branch, as the largest and deepest feature layer of FPN, to produce more robust and fine-grained prototype masks. The concept of prototype masks is similar to those representation learning concepts for object detection from~\cite{sivic2003video, ren2013histograms, agarwal2002learning}. 

We note an important observation here: the learned prototype masks~(feature embeddings) are generalized to different domains. We found that by using different coefficients to linearly assemble the same set of prototype masks, we can generate instance masks and grasp masks.

For the object detection branch, a typical anchor-based object detection branch is extended by adding $N$ extra heads predicting $N \times k$ coefficients for each detected objects. For each object that survives NMS, we predict its class, bounding box, $k$ coefficients for assembling its instance mask, $k$ coefficients for assembling its grasp position mask, $2 \times k$ coefficients for assembling its grasp angle masks~(represented in $sin(2\theta)$ and $cos(2\theta)$, $\theta$ is the valid grasp angle), and $k$ coefficients for assembling its width mask. These predicted sets of coefficients will be used to linearly assemble prototype masks generated from ProtoNet and form target output masks: semantic instance mask, grasp quality mask, grasp angle masks and gripper width mask.

\subsection{Post Processing}
\label{sec:method:post}
\textbf{Target Masks Generation.} As shown in Fig.~\ref{fig:ssg}, the ProtoNet branch will generate $h\times w\times k$ prototype masks $\mathbf{P}$ for the input RGB-D image, where $h, w$ denote the size of the prototype mask. 
$N \times k$ coefficients $\mathbf{C}$ are predicted for every detected object~($N = 5$). Then prototype masks $\mathbf{P}$ are linearly assembled with coefficients $\mathbf{C}$ to generate target masks, $\mathbf{M} = \text{Activation}(\mathbf{P}\mathbf{C}^\top)$. In this study, $\mathbf{M}$ is composed of five masks corresponding to instance mask, grasp quality mask, grasp angle masks and gripper width mask. For instance mask, grasp quality mask and gripper width mask, $Sigmoid$ activation function is used to limit the output range from $0$ to $1$. For grasp angle masks, $tanh$ activation function is used to limit the output range from $-1$ to $1$.

\textbf{Mask Crop.} Generated target masks for each object are cropped using its bounding box. The ground truth bounding boxes are used in training, while the predicted ones are used during evaluation.

\textbf{Grasp Configuration Inference.} The grasp configurations are inferred based on the grasp masks obtained by linearly assembling prototype masks and cropping with bounding boxes. For each object instance, firstly a local maximum point is searched in its grasp quality map to find the point with the highest grasp quality and its pixel coordinates, then the grasp angles and gripper width are obtained from corresponding masks with pixel coordinates of the best grasp point.

\subsection{Loss Function}
\label{sec:method:loss}
Our loss function is composed of five different losses as: object classification~($ \ell_\text{cls}$), bounding box regression~($\ell_{box}$), global semantic segmentation~($\ell_{smask}$), instance segmentation~($\ell_{imask}$), and grasp synthesis~($\ell_{gr}$). $\ell_{cls}$, $\ell_{box}$ and $\ell_{imask}$ are defined the same as in~\cite{bolya2019yolact}. $\ell_{smask}$ is used to accelerate the convergence of our model. $\ell_{g}$ consists of five losses including grasp quality loss~($\ell_{gr-q}$), grasp position loss~($\ell_{gr-p}$), grasp angle loss~(in $sin$ and $cos$, $\ell_{gr-sin}, \ell_{gr-cos}$) and gripper width loss~($\ell_{gr-w}$). $\ell_{gr-q}$, $\ell_{gr-sin}, \ell_{gr-cos}$ and $\ell_{gr-w}$ are calculated using smooth-L1 loss, $\ell_{gr-p}$ is calculated using binary cross entropy loss. The total loss $\mathcal{L}$ is summed as:
\vspace{0mm}
\begin{equation}
\label{eq:loss}
\begin{split} 
\mathcal{L} = &\alpha_{cls}\ell_{cls} + \alpha_{box} \ell_{box} + \alpha_{imask} \ell_{imask} \\
& + \alpha_{gr} \ell_{gr}  + \alpha_{smask} \ell_{smask} 
\end{split},
\end{equation}
\noindent 
where $\ell_{gr} = \alpha_{p}\ell_{gr-p} + \alpha_{q} \ell_{gr-q} + \alpha_{sin}\ell_{gr-sin} + \alpha_{cos}\ell_{gr-cos} + \alpha_{w}\ell_{gr-w}$.

\begin{table*}[t!]

\centering
 
\vspace{-0mm}
\caption{Simulations results: 20 simulated tests were conducted for each object.}
\resizebox{\textwidth}{!}{
\begin{tabular}{|c|c|c|c|c|c|c|c|c|}
\hline
Objects                                                         & Apple  & Banana                                             & Lemon                                               & Mug & Bowl    & Bottle & Marker & \begin{tabular}[c]{@{}c@{}}Cereal Box\end{tabular}    \\ \hline
\begin{tabular}[c]{@{}c@{}}Success Rate {[}\%{]}\end{tabular} & 90     & 85                                                 & 85                                                  & 80  & 75      & 75     & 90     & 90                                                      \\ \hline
Objects                                                         & Sponge & \begin{tabular}[c]{@{}c@{}}Soda Can\end{tabular} & \begin{tabular}[c]{@{}c@{}}Juice Box\end{tabular} & Cup & Spatula & Knife  & Soap   & \begin{tabular}[c]{@{}c@{}}Power Driller\end{tabular} \\ \hline
\begin{tabular}[c]{@{}c@{}}Success Rate {[}\%{]}\end{tabular} & 85     & 90                                                 & 80                                                  & 80  & 80      & 75     & 80     & 85                                                      \\ \hline
\end{tabular}
}
\vspace{-2mm}
\label{tab:sim-exp}
\end{table*}

\begin{figure*}[!htbp]
\centering
\vspace{-2mm}
  \includegraphics[width=0.49\linewidth]{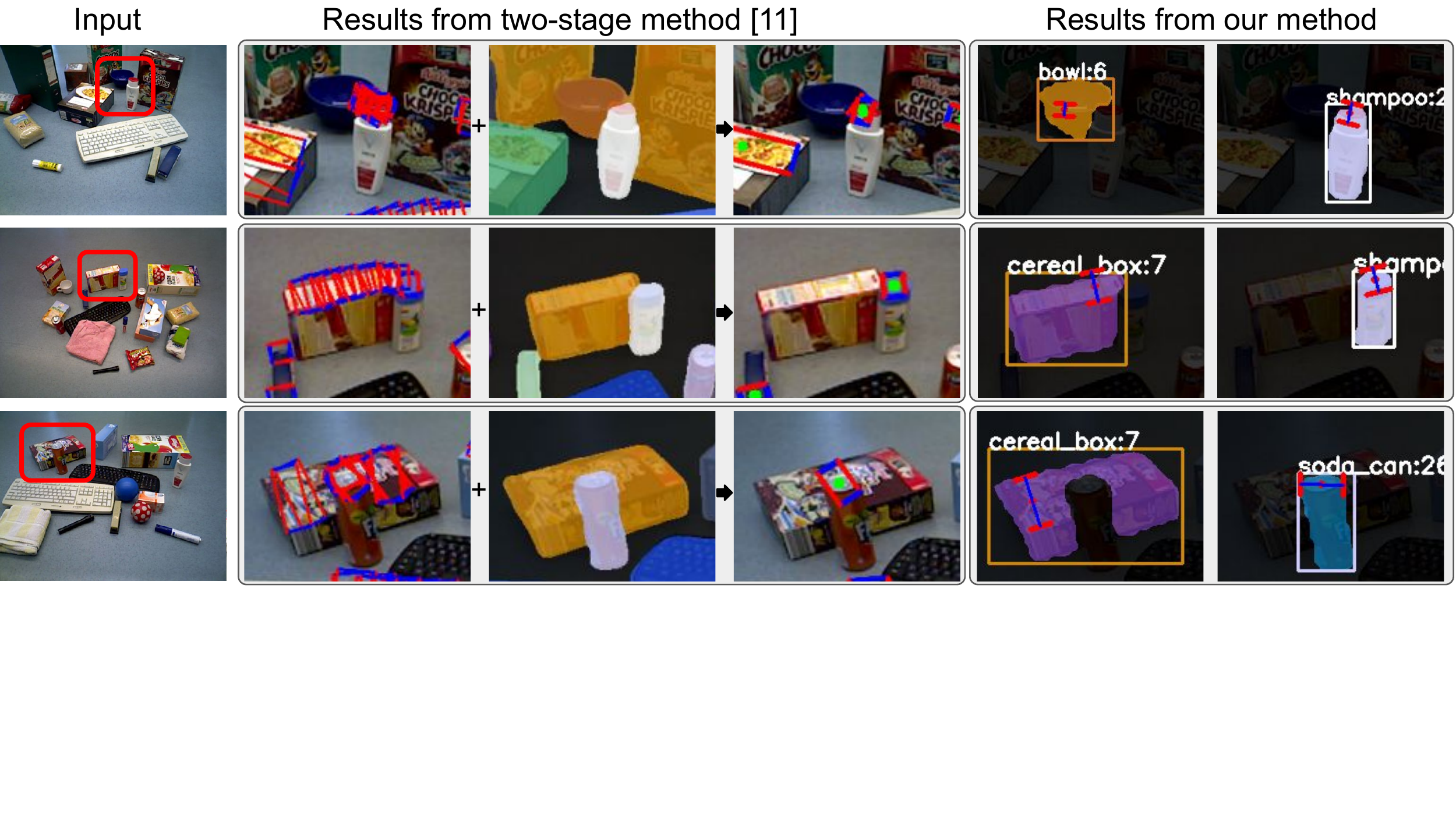}
  \includegraphics[width=0.49\linewidth]{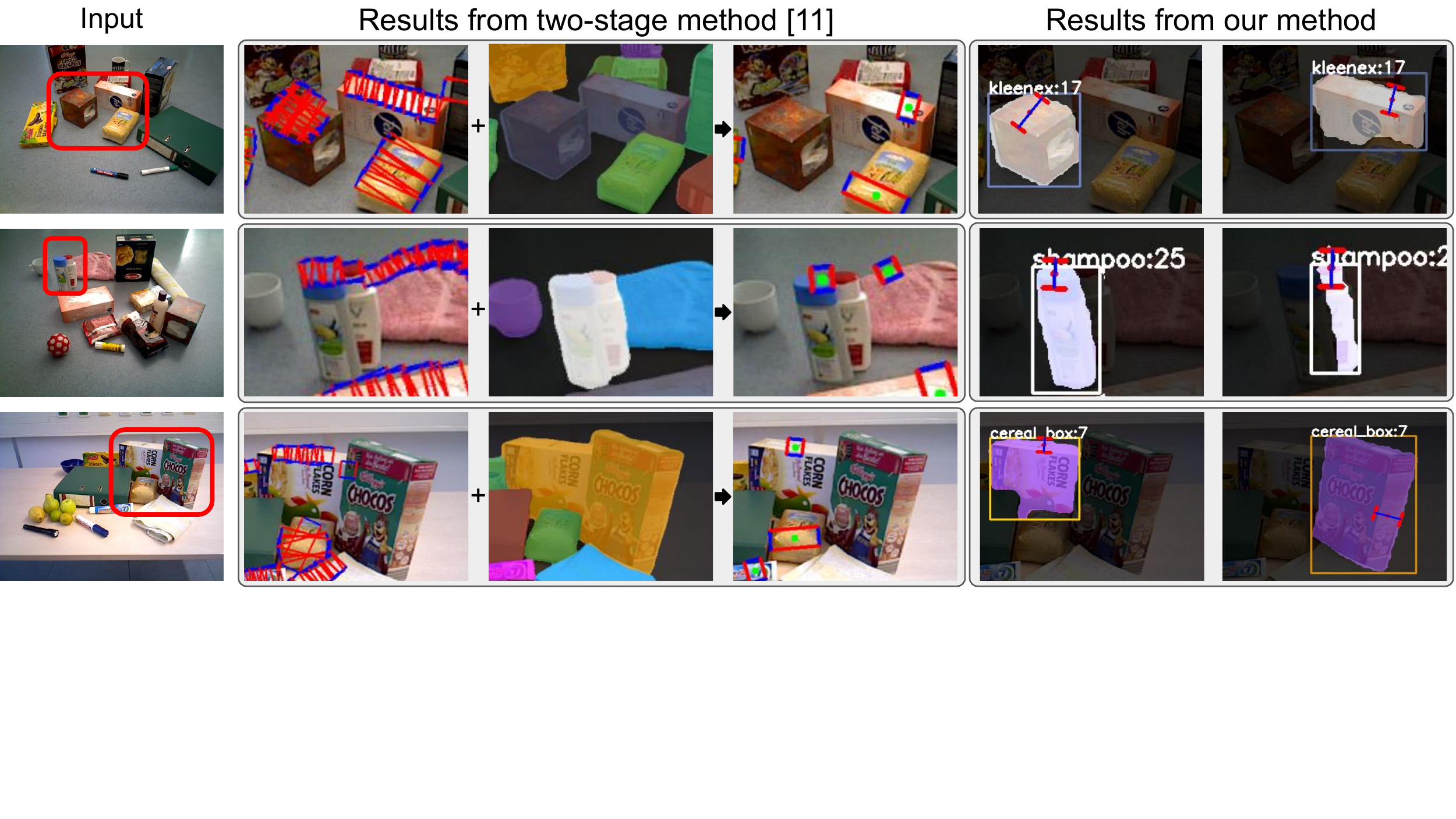}
  \vspace{-15mm}
 
  \caption{Failure cases of two-stage method, compared to the correct results from the proposed single-stage method: (left) Failures of two-stage method caused by inaccurate grasp assignment; (right) Failures of two-stage method caused by segmentation errors.}
\vspace{-5mm}
\label{fig:failure-cases}
\end{figure*}

\section{Evaluation}
\label{sec:evaluation}

We evaluate and benchmark the performance of the proposed method on instance-wise robotic grasp detection dataset OCID-Grasp~\cite{ainetter2021end}, and class-agnostic robotic grasp detection dataset JACQUARD~\cite{depierre2018jacquard}. Moreover, a set of simulations and real robot experiments have been conducted to validate the performance of the proposed method for real-world robotics applications.

To evaluate and quantify the accuracy of predicted grasp configurations of each object on datasets, we applied an extended version of metrics by adding a new condition, based on the Jacquard Index proposed in~\cite{depierre2018jacquard}. That is, a grasp candidate is valid if the following conditions are satisfied: (i) The predicted class label of the grasp candidate is correct; (ii) The angle difference between the predicted grasp candidate and ground truth grasp is within $30^{\circ}$; (iii) The Intersection over Union~(IoU) of the predicted grasp rectangle and the ground truth is greater than $0.25$.

\subsection{Evaluation on OCID-Grasp Dataset}
\label{sec:eval-ocid}

The \textbf{OCID-Grasp dataset} is an extension of  Object Clutter Indoor Dataset~(OCID)~\cite{suchi2019easylabel} annotated by~\cite{ainetter2021end}, which consists of 1763 selected \mbox{RGB-D} images with over 11.4K segmented instance masks and 75K hand-annotated grasp rectangles with corresponding object class information. Objects in OCID-Grasp dataset are classified into 31 different categories. For each scenario, RGB image, depth image, semantic segmentation mask, and grasp annotation with instance labels are provided.

\begin{table}[!t]
\vspace{0mm}
\centering
\caption{Results of grasp accuracy on OCID-Grasp Dataset~\cite{ainetter2021end}.}
\vspace{-2mm}
\resizebox{\linewidth}{!}{
\begin{tabular}{|c|c|c|}
\hline
Method           & Grasp Accuracy  &Speed~(FPS)\\ \hline
Deg\_Seg\_RGB\cite{ainetter2021end}         & 89.02 \%        &31\\ \hline
Deg\_Seg\_RGBD\cite{ainetter2021end}         & 89.84 \%        &31\\ \hline
\textbf{SSG\_RGB}~(ours) & \textbf{91.97} \%         & \textbf{39}\\ \hline
\textbf{SSG\_RGBD}~(ours) & \textbf{92.93} \%         & \textbf{39}\\ \hline
\end{tabular}
}
\label{tab:comparison-ocid-grasp}
\vspace{-4mm}
\end{table}

\begin{figure}[!t]
\centering
\includegraphics[trim=0 8cm 0 0,clip,width=\linewidth]{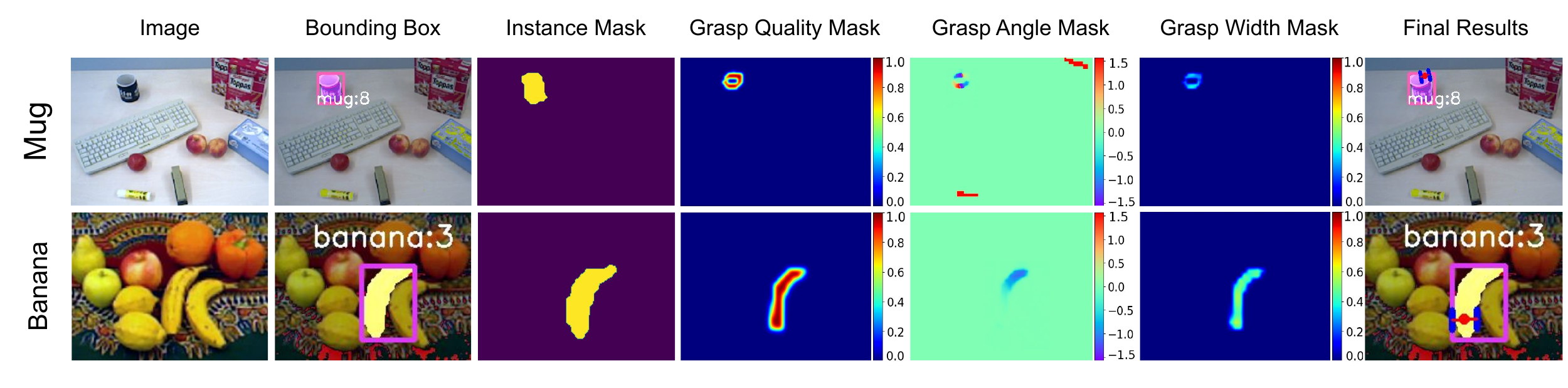}
\vspace{-6mm}
\caption{Test results on OCID-Grasp dataset where grasp configurations were generated for each target object.}
\vspace{-6mm}
\label{fig:ocid-results}
\end{figure}

On OCID-Grasp dataset, our model is trained on the official train set and validated on the test set. To augment the size of datasets for training our network, based on OCID-Grasp, we applied random data augmentation including random photometric distortion, random clip and multi-scale resize. We outperform state-of-the-art on OCID-Grasp dataset with an overall grasp accuracy of $92.9\%$. The results are summarised in TABLE~\ref{tab:comparison-ocid-grasp}.

In comparison with the two-stage baseline method as in~\cite{ainetter2021end}, our method perform instance segmentation and instance-wise grasp synthesis \textit{simultaneously} to synthesize grasp configuration in a single stage, with the accuracy of $92.9\%$ and the inference speed of $39$ frames per second -- which has outperformed the baseline with a significant margin by $3.91\%$ in accuracy and $25\%$ in inference speed. Fig.~\ref{fig:ocid-results} shows the results of two representative test samples from the OCID-Grasp dataset. To better support the advance of our proposed method, representative cases are show in Fig.~\ref{fig:failure-cases} in which two-stage method~\cite{ainetter2021end}) failed while our proposed method succeeded.

\begin{table}[!htbp]
\centering

\caption{Comparison of grasp accuracy ([\%]) on JACQUARD Dataset \cite{depierre2018jacquard}, with different IoU thresholds and angle threshold of $30^{\circ}$. Results for~\cite{ainetter2021end, zhou2018fully, depierre2021scoring} are taken from~\cite{ainetter2021end}. Results of~\cite{kumra2020antipodal} are reproduced.}
\vspace{-2mm}
\resizebox{\linewidth}{!}{\begin{tabular}{|c|c|c|c|}
\hline
Method                     & IoU 25\%        & IoU 30\%         & IoU 35\%         \\ \hline
Method of ~\cite{zhou2018fully}                       & 81.95         & 78.26          & 74.33          \\ \hline
Method of ~\cite{depierre2021scoring}                   & 85.74         & 82.58          & 78.71          \\ \hline
Mask-Grasp RCNN~\cite{kamel2021mask}                      & 89.80  & -       & -             \\ \hline
Method of ~\cite{SONG2020101963}                       & 91.5          & 89.7           & 87.3           \\ \hline
Gr-ConvNet~\cite{kumra2020antipodal}                      & 91.83          & 89.55                & 85.99                \\ \hline

Det~\cite{ainetter2021end}              & 92.69         & 91.29          & 88.99          \\ \hline
Det-Seg-Refine~\cite{ainetter2021end} & 92.95         & 91.33          & 88.96          \\ \hline
\textbf{SSG}(Ours)              & \textbf{91.8} & \textbf{89.95} & \textbf{88.49} \\ \hline
\end{tabular}}
\label{tab:comparison-jacquard-iou}
\vspace{-4mm}
\end{table}

\begin{table}[!htbp]
\centering
\caption{Comparison of grasp accuracy~([\%]) for JACQUARD Dataset~\cite{depierre2018jacquard}, with different angle thresholds and IoU threshold of 25\%. Results of~\cite{ainetter2021end, zhou2018fully, depierre2021scoring} are referenced  from~\cite{ainetter2021end}, and results of \cite{kumra2020antipodal} are reproduced.}
\vspace{-2mm}
\resizebox{\linewidth}{!}{\begin{tabular}{|c|c|c|c|c|c|c|}
\hline
Method                     & $30^{\circ}$      & $25^{\circ}$      & $20^{\circ}$      & $15^{\circ}$      & $10^{\circ}$      & $5^{\circ}$       \\ \hline
Method of ~\cite{zhou2018fully}                       & 81.95 & 81.76 & 81.27 & 80.23 & 77.79 & -       \\ \hline
Method of ~\cite{depierre2021scoring}                   & 85.74 & 85.55 & 85.01 & 83.65 & 80.82 & -       \\ \hline
Mask-Grasp RCNN~\cite{kamel2021mask}                      & 89.80  & -       & -       & -       & -       & -       \\ \hline
Gr-ConvNet~\cite{kumra2020antipodal}                      & 91.83  & 90.00       & 87.34       & 83.45       & 77.94       & 63.67       \\ \hline
Det~\cite{ainetter2021end}              & 92.68 & 92.34 & 92.08 & 91.40 & 88.12 & 56.23 \\ \hline
Det-Seg-Refine~\cite{ainetter2021end} & 92.95 & 92.88 & 92.42 & 91.52 & 88.12 & 72.79 \\ \hline
\textbf{SSG}(Ours)                       & \textbf{91.8} & \textbf{91.11} & \textbf{90.05} & \textbf{87.97} & \textbf{81.68} & \textbf{60.87} \\ \hline
\end{tabular}}
\label{tab:comparison-jacquard-angle}
\vspace{-4mm}
\end{table}

\vspace{-0mm}

\subsection{Evaluation on JACQUARD Dataset}
\label{sec:eval-jacquard}

The \textbf{JACQUARD Dataset} is built on a subset of ShapeNet~\cite{shapenet2015} which is a large CAD model dataset. It consists of 54485 different scenes from 11619 distinct objects. In total, it has over 4.9M grasp annotations~(from over 1.1M unique locations). For each scenario, a render RGB image, a segmentation mask, two depth images and grasp annotations are provided.

However, the JACQUARD Dataset only contains single-object scenes without class labels for grasp annotations. Thus, we applied minimal adaptation of our method and make it a class-agnostic one, we labeled all objects as ``object''. Our model was trained on JACQUARD dataset in a class-agnostic way and evaluated using several metrics with different thresholds. Detailed results are summarised in TABLE~\ref{tab:comparison-jacquard-angle} and TABLE~\ref{tab:comparison-jacquard-iou}~(Unavailable results were denoted as ``-'').

\begin{figure}[t]
\centering
\vspace{-0mm}
  \includegraphics[width=0.35\linewidth]{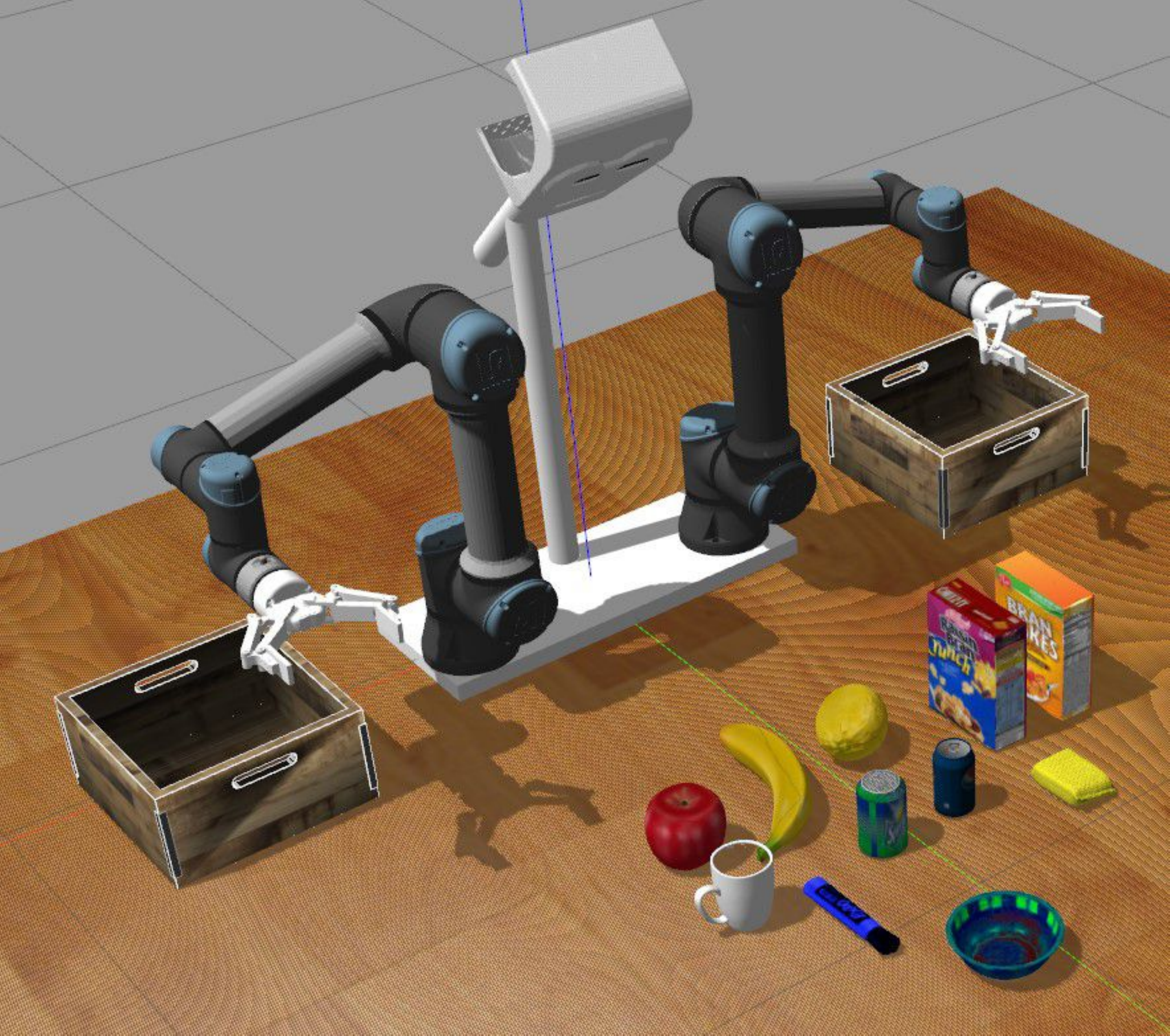}
  \includegraphics[width=0.35\linewidth]{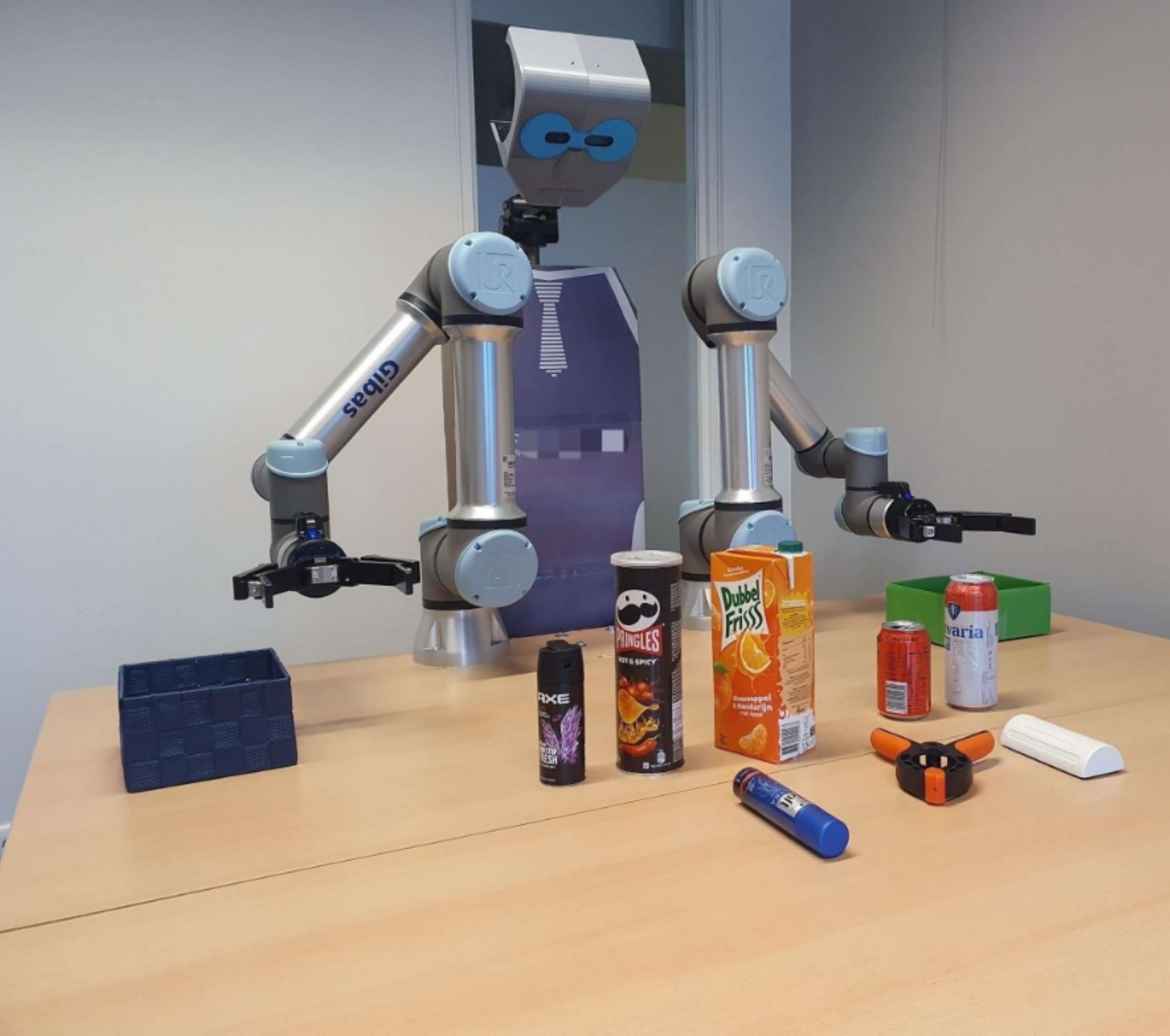}
 
  \caption{Simulation and experiment setups.}
\vspace{-5mm}
\label{fig:simExpSetup}
\end{figure}

The evaluation on the JACQUARD Dataset show that our method is generalized and can predict both high-quality instance masks and grasp masks for general objects without class-specific information. Despite the lack of class-specific information, our approach was very competitive among learning-based 2D grasp synthesis approaches. Further, we have conducted ablation study to support the importance of class-specific information (Details in Section~\ref{tab:ablation-study}). Replacing the detection and segmentation heads with a class-agnostic one, such like~\cite{xie2021unseen}, could be a potential way to boost the performance of our method on the JACQUARD Dataset. 

We note that our proposed method, the SSG, significantly surpasses the Mask-Grasp RCNN~\cite{kamel2021mask} which is another single-stage instance-wise grasp synthesis method based on Mask-RCNN~\cite{he2017mask}. Our method has reached 91.8\% grasp accuracy on the Jacquard dataset~\cite{depierre2018jacquard} which outperforms the Mask-Grasp RCNN~\cite{kamel2021mask} by 2\%. Moreover, our method can run inference at 39FPS rate, which is almost three times faster than the Mask-Grasp RCNN~\cite{kamel2021mask}~(14FPS).

\begin{figure}[t]
\centering
\includegraphics[trim=0 0.6cm 3cm 0cm,clip,width=0.8\linewidth]{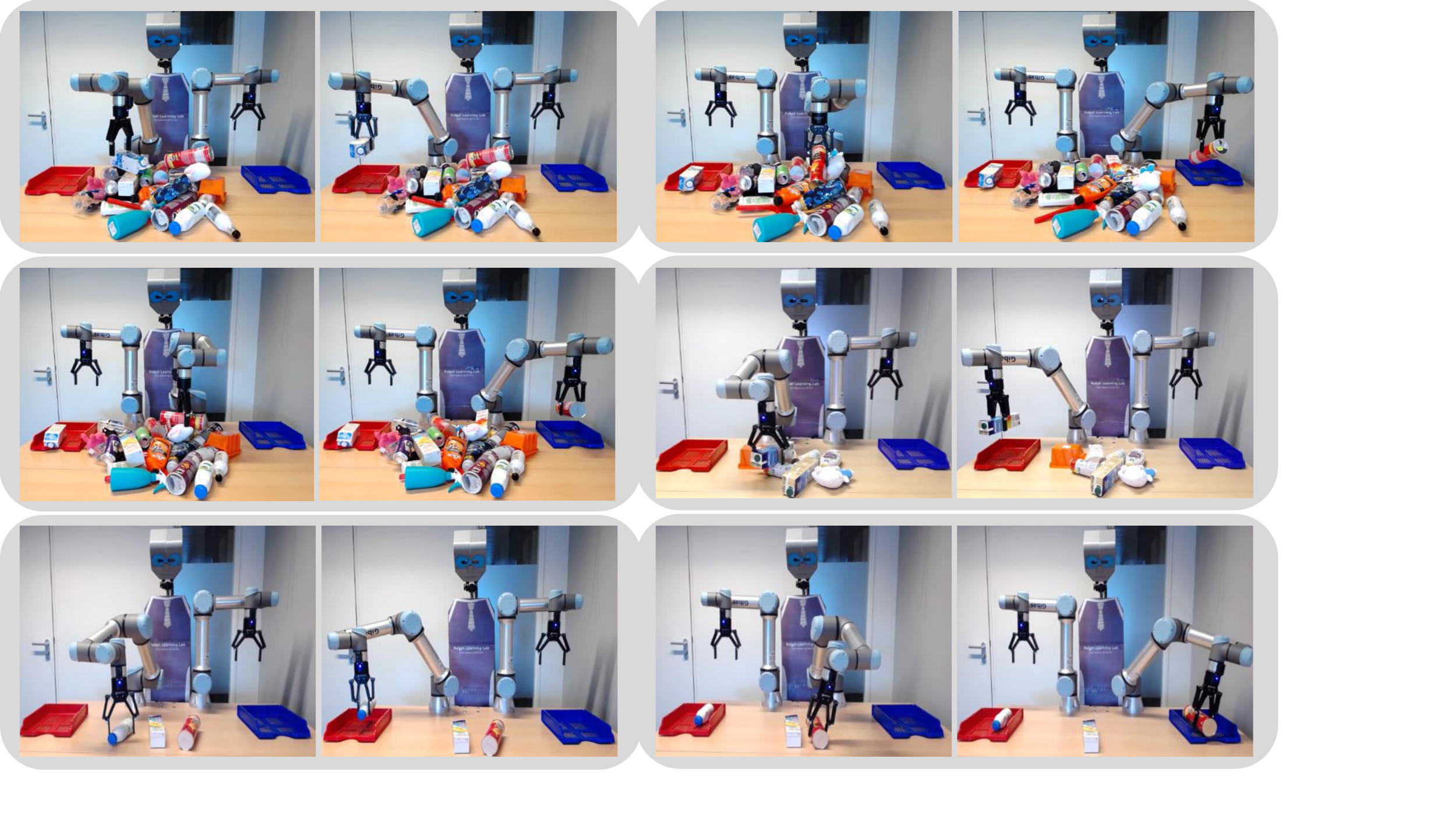}
\vspace{-2mm}
\caption{Real robot performing the table cleaning task in three different levels of difficulties: highly cluttered, cluttered and isolated real-world scenarios.}
\label{fig:realRobot}
\vspace{-7mm}
\end{figure}

\begin{table*}[t]
\centering
\caption{Ablation study on OCID-Grasp Dataset~\cite{ainetter2021end}.}
\resizebox{0.95\linewidth}{!}{
\begin{tabular}{|c|c|c|c|c|c|c|}
\hline
Model            & SSG   & \begin{tabular}[c]{@{}c@{}}SSG\\ without\\ instace segmentation\end{tabular} & \begin{tabular}[c]{@{}c@{}}SSG\\ without\\ class prediction\end{tabular} & SSG       & \begin{tabular}[c]{@{}c@{}}SSG\\ without\\ instance segmentation\end{tabular} & \begin{tabular}[c]{@{}c@{}}SSG\\ without\\ class prediction\end{tabular} \\ \hline
Input Modalities & RGB   & RGB                                                                          & RGB                                                                      & RGB+Depth & RGB+Depth                                                                     & RGB+Depth                                                                \\ \hline
Grasp Accuracy   & 91.97\% & 90.92\%                                                                        & 90.31\%                                                                    & 92.93\%     & 92.09\%                                                                         & 90.81\%                                                                    \\ \hline
\end{tabular}}
\vspace{-3mm}
\label{tab:ablation-study}
\end{table*}

\subsection{Simulation and Real Robot Experiments}

A set of simulations and real robot experiments have been conducted to validate that our model can be used to generate high-quality grasp candidates for robotic manipulators: (i)~single object pick-and-place task in simulation; (ii)~table cleaning task using a real robot. 

Our simulations and experiments focused on table top domains, where objects are in arbitrary spatial arrangements on the table. The simulation setup used a synthetic dataset from~\cite{kasaei2022lifelong} which contains 90 simulated house-hold objects, imported from different resources, e.g., YCB dataset~\cite{sener2017active}, Gazebo repository. The whole setup is composed of a dual arm robot with two UR5 manipulators and a Kinect sensor to acquire RGB-D images. For real robot experiments, we used exactly the same setup (see Fig.~\ref{fig:simExpSetup}). We trained a model using OCID-Grasp dataset~\cite{ainetter2021end} which is used for both simulations and experiments.

In the first task, 16 objects, including 6 unseen objects~(Juice Box, Cup, Spatula, Knife, Soap and Power Driller) have been selected. In each trial, one of them was randomly put on a table for 20 rounds of pick-and-place. A grasp configuration is considered successful if the object can be grasped, lifted up and placed at the designed place. The success rate for each object has been summarized in TABLE~\ref{tab:sim-exp}.


The second task is focused on validating the proposed method on a real robot for table cleaning. In this task, an operator randomly places a set of unseen objects on the table and the robot should remove and place them into the predefined targets one by one. This task has been repeated in 3 different levels of difficulties: isolated (less than 3 objects), cluttered~(less than 10 objects) and highly cluttered~(more than 15 objects). A set of snapshots is shown in Fig.~\ref{fig:realRobot}. We performed ten rounds of experiments per level, and assessed the performance by the success rate, where the attempt is considered as a success if the target object can be grasped and moved to the target successfully. The results showed that the robot is able to accomplish the task with success rates of $84.0\%$, $79.4\%$ and $71.3\%$, respectively. It should be noted that in some failure cases in (highly) cluttered environments, although the grasp predictions were correct, execution of grasps were not feasible due to either the limitation of the motion planning or prevention of the grasp action in presence of the surrounding objects, rather than due to the grasp predictions themselves. The video of our experiments is available at \url{https://youtu.be/riBXMgrupUw.}

\begin{figure}[t]
\centering
\vspace{-3mm}
\includegraphics[trim=0 5.8cm 0 0,clip,width=\linewidth]{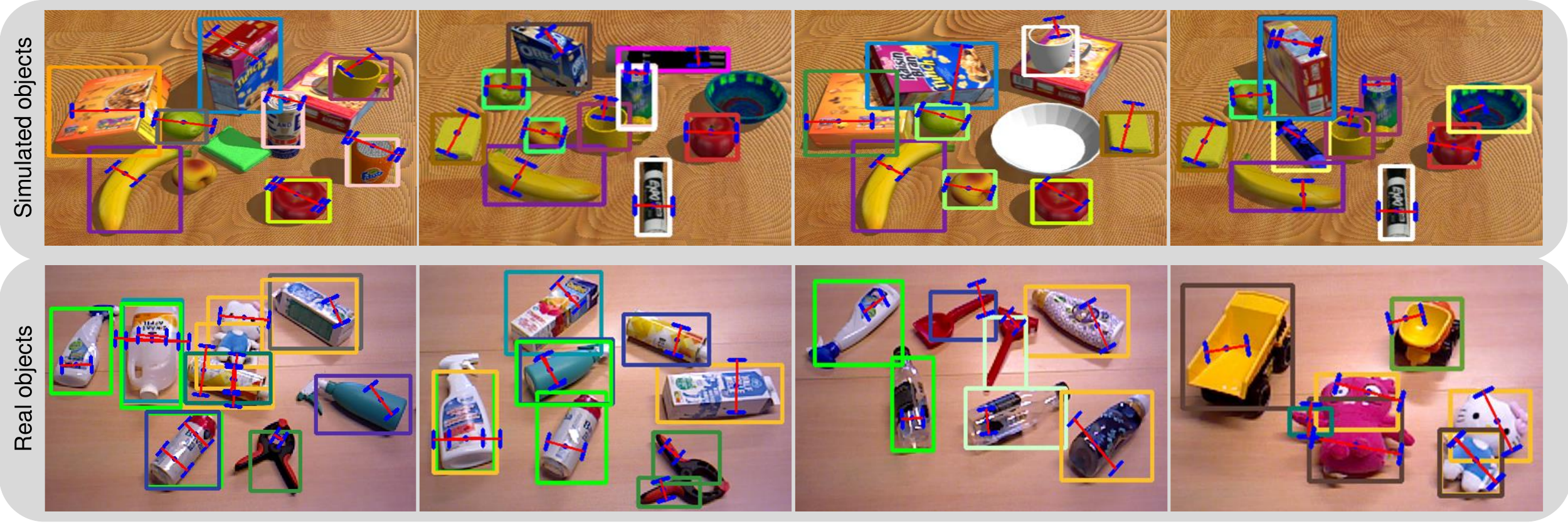}
\caption{Results of a set of tests in clutter environments on simulated and real objects.}
\vspace{-6mm}
\label{fig:unseen}
\end{figure}

\subsection{Scalability}
\label{sec:scalability}

The success of our proposed SSG shows the potential and scalability of feature assembling using linear coefficients. To further prove the scalability of our method, 
we re-train our model on Object Stacking Grasping Dataset~(OSGD)~\cite{zhang2019multi}, which includes additional affordance annotations from total 11 different types of grasping actions including cut, write, hammer, fork, shovel, wrench, pinch, screw, ladle, brush and hand-over. Here, the affordance annotation refers to the correct grasping action~(e.g., knife -- cut, screwdriver -- screw, etc.). 

For each object sample from OSGD dataset, its class label, bounding box, grasp annotations~(in rectangles) and affordance annotations are provided. 
To generate additional affordance masks for each detected object, we add 11 extra heads in the object detection branch to predict 11 sets of coefficients to linearly assemble the shared prototype masks, and generate 11 target affordance masks. Since this dataset does not provide the instance semantic masks, our architecture is adopted accordingly: the global semantic segmentation head and the instance mask head are removed. 
Moreover, the OSGD dataset only provides depth image as input, thus the input channel of the feature extraction module is changed and no pre-trained model is used to initialize the feature extraction module. 

\begin{figure}[t]
\vspace{-4mm}
\centering
  \includegraphics[trim=1cm 0 1cm 0,clip,width=0.9\linewidth]{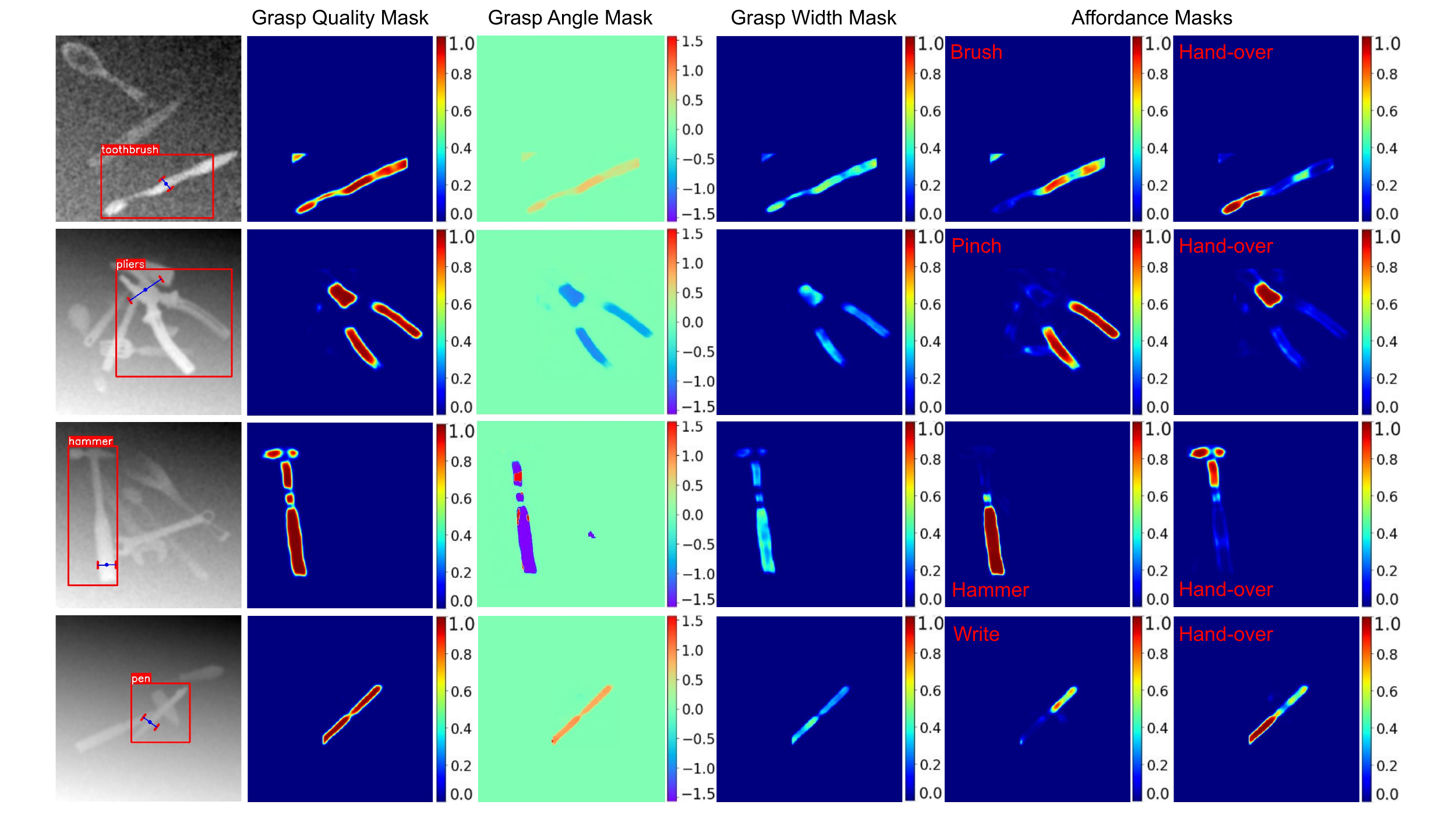}
  \smallskip\par
  \vspace{-4mm}
  \caption{Validation results on the OSGD dataset~\cite{zhang2019multi}, further showing our method can generate additional \textit{affordance masks} by predicting more sets of coefficients and assembling prototype masks with these coefficients.}
\label{fig:osgd-results}
\vspace{-6mm}
\end{figure}

As is shown in Fig.~\ref{fig:osgd-results}, correct affordance masks as well as grasp configurations are generated for different target objects which prove the scalability and extendability of the our proposed model. It can be extended to predict extra target masks by simply adding more coefficients predicting heads without changing the overall complexity. This feature of our methodology, in our opinion, will have an great impact on the field of robotic grasping synthesis research.

\subsection{Ablation Study}
\label{sec:method:ablation}
A set of ablation study was conducted to support the current design. The detection head of our model is composed of three modules: object detection, instance segmentation, and generation of grasp maps. To validate the proposed network design, we have retrained and tested two additional models on the OCID-Grasp dataset~\cite{ainetter2021end}: (1) a model without predicting object class label; (2) a model without generating instance mask. The detailed results are shown in the TABLE~\ref{tab:ablation-study}.

It can be seen from TABLE~\ref{tab:ablation-study} that the depth channel brings useful information and increases the performance. The instance segmentation module and the class prediction module also play an important role for grasp synthesis. Without instance segmentation head, the overall grasp accuracy of our model on OCID-Grasp dataset~\cite{ainetter2021end} decreases from 91.97\% to 90.92\% (with RGB input), and from 92.93\% to 92.09\% (with RGB-D input). Without class prediction head, the overall grasp accuracy of our model on OCID-Grasp dataset~\cite{ainetter2021end} decreases from 91.97\% to 90.31\% (with RGB input), and from 92.93\% to 90.81\% (with RGB-D input).  The results of the ablation study has shown the benefits of our proposed network design: Generating different target masks by linearly assembling the same set of learned feature maps with different coefficients, which is able to exchange features across different domains, and also to make the learned feature maps more general and robust.

\vspace{-5mm}
\section{Conclusion}
\label{sec:limitation}
This work developed a novel single-stage grasp synthesis model -- SSG -- for tackling instance-wise grasp synthesis task in a single-stage manner. Our method formulated the instance-wise grasp synthesis as two sub-tasks: first, a set of learned feature embeddings is generated, which captures general features of the input RGB-D image; second, anchor-based object detection is conducted. For each detection, five sets of coefficients are predicted that will be used to linearly assemble generated feature embeddings to form a semantic instance mask and four grasp masks, simultaneously. We evaluated our method on the well-known JACQUARD dataset and a more challenging OCID-Grasp dataset. The results showed that our method outperforms the state-of-the-art on OCID-Grasp dataset and performs competitively on JACQUARD dataset. Moreover, the proposed method has been extensively tested both in simulation and on the real robot, using isolated, cluttered and highly cluttered scenarios. All these extensive results validated that our method can generate valid grasp configurations for target objects in multi-object scenarios.





\bibliographystyle{IEEEtran}
\bibliography{SSG-Grasping}

\begin{thebibliography}{10}
\providecommand{\url}[1]{#1}
\csname url@rmstyle\endcsname
\providecommand{\newblock}{\relax}
\providecommand{\bibinfo}[2]{#2}
\providecommand\BIBentrySTDinterwordspacing{\spaceskip=0pt\relax}
\providecommand\BIBentryALTinterwordstretchfactor{4}
\providecommand\BIBentryALTinterwordspacing{\spaceskip=\fontdimen2\font plus
\BIBentryALTinterwordstretchfactor\fontdimen3\font minus
  \fontdimen4\font\relax}
\providecommand\BIBforeignlanguage[2]{{%
\expandafter\ifx\csname l@#1\endcsname\relax
\typeout{** WARNING: IEEEtran.bst: No hyphenation pattern has been}%
\typeout{** loaded for the language `#1'. Using the pattern for}%
\typeout{** the default language instead.}%
\else
\language=\csname l@#1\endcsname
\fi
#2}}

\bibitem{8793751}
H.~Karaoguz and P.~Jensfelt, ``Object detection approach for robot grasp
  detection,'' in \emph{2019 International Conference on Robotics and
  Automation (ICRA)}, 2019, pp. 4953--4959.

\bibitem{SONG2020101963}
\BIBentryALTinterwordspacing
Y.~Song, L.~Gao, X.~Li, and W.~Shen, ``A novel robotic grasp detection method
  based on region proposal networks,'' \emph{Robotics and Computer-Integrated
  Manufacturing}, vol.~65, p. 101963, 2020. [Online]. Available:
  \url{https://www.sciencedirect.com/science/article/pii/S0736584519308105}
\BIBentrySTDinterwordspacing

\bibitem{zhang2019roi}
H.~Zhang, X.~Lan, S.~Bai, X.~Zhou, Z.~Tian, and N.~Zheng, ``Roi-based robotic
  grasp detection for object overlapping scenes,'' in \emph{2019 IEEE/RSJ
  International Conference on Intelligent Robots and Systems (IROS)}.\hskip 1em
  plus 0.5em minus 0.4em\relax IEEE, 2019, pp. 4768--4775.

\bibitem{luo2020grasp}
Z.~Luo, B.~Tang, S.~Jiang, M.~Pang, and K.~Xiang, ``Grasp detection based on
  faster region cnn,'' in \emph{2020 5th International Conference on Advanced
  Robotics and Mechatronics (ICARM)}.\hskip 1em plus 0.5em minus 0.4em\relax
  IEEE, 2020, pp. 323--328.

\bibitem{morrison2018closing}
D.~Morrison, P.~Corke, and J.~Leitner, ``Closing the loop for robotic grasping:
  A real-time, generative grasp synthesis approach,'' \emph{arXiv preprint
  arXiv:1804.05172}, 2018.

\bibitem{kumra2020antipodal}
S.~Kumra, S.~Joshi, and F.~Sahin, ``Antipodal robotic grasping using generative
  residual convolutional neural network,'' in \emph{2020 IEEE/RSJ International
  Conference on Intelligent Robots and Systems (IROS)}.\hskip 1em plus 0.5em
  minus 0.4em\relax IEEE, 2020, pp. 9626--9633.

\bibitem{cao2021lightweight}
H.~Cao, G.~Chen, Z.~Li, J.~Lin, and A.~Knoll, ``Lightweight convolutional
  neural network with gaussian-based grasping representation for robotic
  grasping detection,'' \emph{arXiv preprint arXiv:2101.10226}, 2021.

\bibitem{li2022novel}
Y.~Li, Y.~Liu, Z.~Ma, and P.~Huang, ``A novel generative convolutional neural
  network for robot grasp detection on gaussian guidance,'' \emph{arXiv
  preprint arXiv:2205.04003}, 2022.

\bibitem{zhang2019multi}
H.~Zhang, X.~Lan, S.~Bai, L.~Wan, C.~Yang, and N.~Zheng, ``A multi-task
  convolutional neural network for autonomous robotic grasping in object
  stacking scenes,'' in \emph{2019 IEEE/RSJ International Conference on
  Intelligent Robots and Systems (IROS)}.\hskip 1em plus 0.5em minus
  0.4em\relax IEEE, 2019, pp. 6435--6442.

\bibitem{yang2019task}
C.~Yang, X.~Lan, H.~Zhang, and N.~Zheng, ``Task-oriented grasping in object
  stacking scenes with crf-based semantic model,'' in \emph{2019 IEEE/RSJ
  International Conference on Intelligent Robots and Systems (IROS)}.\hskip 1em
  plus 0.5em minus 0.4em\relax IEEE, 2019, pp. 6427--6434.

\bibitem{ainetter2021end}
S.~Ainetter and F.~Fraundorfer, ``End-to-end trainable deep neural network for
  robotic grasp detection and semantic segmentation from rgb,'' in \emph{2021
  IEEE International Conference on Robotics and Automation (ICRA)}.\hskip 1em
  plus 0.5em minus 0.4em\relax IEEE, 2021, pp. 13\,452--13\,458.

\bibitem{li2021keypoint}
T.~Li, F.~Wang, C.~Ru, Y.~Jiang, and J.~Li, ``Keypoint-based robotic grasp
  detection scheme in multi-object scenes,'' \emph{Sensors}, vol.~21, no.~6, p.
  2132, 2021.

\bibitem{bohg2013data}
J.~Bohg, A.~Morales, T.~Asfour, and D.~Kragic, ``Data-driven grasp
  synthesis—a survey,'' \emph{IEEE Transactions on robotics}, vol.~30, no.~2,
  pp. 289--309, 2013.

\bibitem{lenz2015deep}
I.~Lenz, H.~Lee, and A.~Saxena, ``Deep learning for detecting robotic grasps,''
  \emph{The International Journal of Robotics Research}, vol.~34, no. 4-5, pp.
  705--724, 2015.

\bibitem{redmon2015real}
J.~Redmon and A.~Angelova, ``Real-time grasp detection using convolutional
  neural networks,'' in \emph{2015 IEEE international conference on robotics
  and automation (ICRA)}.\hskip 1em plus 0.5em minus 0.4em\relax IEEE, 2015,
  pp. 1316--1322.

\bibitem{cheng2022robot}
H.~Cheng, Y.~Wang, and M.~Q.-H. Meng, ``A robot grasping system with
  single-stage anchor-free deep grasp detector,'' \emph{IEEE Transactions on
  Instrumentation and Measurement}, vol.~71, pp. 1--12, 2022.

\bibitem{kamel2021mask}
M.~S. Kamel and M.~D. Naish, ``Mask-grasp r-cnn: Simultaneous instance
  segmentation and robotic grasp detection,'' in \emph{2021 IEEE EMBS
  International Conference on Biomedical and Health Informatics (BHI)}.\hskip
  1em plus 0.5em minus 0.4em\relax IEEE, 2021, pp. 1--6.

\bibitem{he2017mask}
K.~He, G.~Gkioxari, P.~Doll{\'a}r, and R.~Girshick, ``Mask r-cnn,'' in
  \emph{Proceedings of the IEEE international conference on computer vision},
  2017, pp. 2961--2969.

\bibitem{bolya2019yolact}
D.~Bolya, C.~Zhou, F.~Xiao, and Y.~J. Lee, ``Yolact: Real-time instance
  segmentation,'' in \emph{Proceedings of the IEEE/CVF international conference
  on computer vision}, 2019, pp. 9157--9166.

\bibitem{he2016deep}
K.~He, X.~Zhang, S.~Ren, and J.~Sun, ``Deep residual learning for image
  recognition,'' in \emph{Proceedings of the IEEE conference on computer vision
  and pattern recognition}, 2016, pp. 770--778.

\bibitem{lin2017feature}
T.-Y. Lin, P.~Doll{\'a}r, R.~Girshick, K.~He, B.~Hariharan, and S.~Belongie,
  ``Feature pyramid networks for object detection,'' in \emph{Proceedings of
  the IEEE conference on computer vision and pattern recognition}, 2017, pp.
  2117--2125.

\bibitem{long2015fully}
J.~Long, E.~Shelhamer, and T.~Darrell, ``Fully convolutional networks for
  semantic segmentation,'' in \emph{Proceedings of the IEEE conference on
  computer vision and pattern recognition}, 2015, pp. 3431--3440.

\bibitem{sivic2003video}
J.~Sivic and A.~Zisserman, ``Video google: A text retrieval approach to object
  matching in videos,'' in \emph{Computer Vision, IEEE International Conference
  on}, vol.~3.\hskip 1em plus 0.5em minus 0.4em\relax IEEE Computer Society,
  2003, pp. 1470--1470.

\bibitem{ren2013histograms}
X.~Ren and D.~Ramanan, ``Histograms of sparse codes for object detection,'' in
  \emph{Proceedings of the IEEE conference on computer vision and pattern
  recognition}, 2013, pp. 3246--3253.

\bibitem{agarwal2002learning}
S.~Agarwal and D.~Roth, ``Learning a sparse representation for object
  detection,'' in \emph{European conference on computer vision}.\hskip 1em plus
  0.5em minus 0.4em\relax Springer, 2002, pp. 113--127.

\bibitem{depierre2018jacquard}
A.~Depierre, E.~Dellandr{\'e}a, and L.~Chen, ``Jacquard: A large scale dataset
  for robotic grasp detection,'' in \emph{2018 IEEE/RSJ International
  Conference on Intelligent Robots and Systems (IROS)}.\hskip 1em plus 0.5em
  minus 0.4em\relax IEEE, 2018, pp. 3511--3516.

\bibitem{suchi2019easylabel}
M.~Suchi, T.~Patten, D.~Fischinger, and M.~Vincze, ``Easylabel: A
  semi-automatic pixel-wise object annotation tool for creating robotic rgb-d
  datasets,'' in \emph{2019 International Conference on Robotics and Automation
  (ICRA)}.\hskip 1em plus 0.5em minus 0.4em\relax IEEE, 2019, pp. 6678--6684.

\bibitem{zhou2018fully}
X.~Zhou, X.~Lan, H.~Zhang, Z.~Tian, Y.~Zhang, and N.~Zheng, ``Fully
  convolutional grasp detection network with oriented anchor box,'' in
  \emph{2018 IEEE/RSJ International Conference on Intelligent Robots and
  Systems (IROS)}.\hskip 1em plus 0.5em minus 0.4em\relax IEEE, 2018, pp.
  7223--7230.

\bibitem{depierre2021scoring}
A.~Depierre, E.~Dellandr{\'e}a, and L.~Chen, ``Scoring graspability based on
  grasp regression for better grasp prediction,'' in \emph{2021 IEEE
  International Conference on Robotics and Automation (ICRA)}.\hskip 1em plus
  0.5em minus 0.4em\relax IEEE, 2021, pp. 4370--4376.

\bibitem{shapenet2015}
A.~X. Chang, T.~Funkhouser, L.~Guibas, P.~Hanrahan, Q.~Huang, Z.~Li,
  S.~Savarese, M.~Savva, S.~Song, H.~Su, J.~Xiao, L.~Yi, and F.~Yu,
  ``{ShapeNet: An Information-Rich 3D Model Repository},'' Stanford University
  --- Princeton University --- Toyota Technological Institute at Chicago, Tech.
  Rep. arXiv:1512.03012 [cs.GR], 2015.

\bibitem{xie2021unseen}
C.~Xie, Y.~Xiang, A.~Mousavian, and D.~Fox, ``Unseen object instance
  segmentation for robotic environments,'' \emph{IEEE Transactions on
  Robotics}, vol.~37, no.~5, pp. 1343--1359, 2021.

\bibitem{kasaei2022lifelong}
H.~Kasaei and S.~Xiong, ``Lifelong ensemble learning based on multiple
  representations for few-shot object recognition,'' \emph{arXiv preprint
  arXiv:2205.01982}, 2022.

\bibitem{sener2017active}
O.~Sener and S.~Savarese, ``Active learning for convolutional neural networks:
  A core-set approach,'' \emph{arXiv preprint arXiv:1708.00489}, 2017.

\end{thebibliography}
\end{document}